\documentclass[double]{IEEEtran}

\usepackage{latexsym,,amssymb,amsmath,graphicx,epsf,cite,bbm,float}
\usepackage{ifpdf}
\usepackage{epstopdf}

\usepackage{algorithm,algorithmic}
\usepackage{amsmath,amssymb,bm}
\usepackage{amsfonts,dsfont,color,bbm,subcaption}

\usepackage{amsmath,amsthm}
\newcommand{\be}{\begin{equation}}
\newcommand{\ee}{\end{equation}}
\newcommand{\bea}{\begin{eqnarray}}
\newcommand{\eea}{\end{eqnarray}}

\newcommand{\MB}{\left[\begin{array}}
\newcommand{\ME}{\end{array}\right]}

\newcommand{\ei}{\end{itemize}}
\newcommand{\bi}{\begin{itemize}}

\newcommand{\xt}[1][t]{x_{#1}}

\newcommand{\lt}[1][t]{l_{#1}}
\newcommand{\ft}[1][t]{f_{#1}}
\newcommand{\pt}[1][i,t]{p_{#1}}
\newcommand{\tpt}[1][i,t]{\widetilde{p}_{#1}}
\newcommand{\gt}[1]{\gamma_{#1}}
\newcommand{\expect}[1]{\mathbb{E}\left[#1\right]}
\newcommand{\xxt}[1][t]{{\xt[#1]}^{(1)}}
\newcommand{\ppt}[1][i,t]{\pt[#1]^{(1)}}
\newcommand{\tppt}[1][i,t]{\widetilde{p}_{#1}^{(1)}}
\newcommand{\xxxt}[1][t]{{\xt[t]}^{(2)}}

\DeclareMathOperator*{\argmin}{arg\,min}

\newcommand{\tO}{\widetilde{O}}

\usepackage{mathtools}
\DeclarePairedDelimiter\ceil{\lceil}{\rceil}

\newtheorem{theorem}{Theorem}
\newtheorem{definition}[]{Definition}
\newtheorem{assumption}[]{Assumption}
\newtheorem{lemma}[]{Lemma}
\newtheorem{corollary}[]{Corollary}
\newtheorem{remark}[]{Remark}
\begin{document}

\title{Recursive Experts: An Efficient Optimal Mixture of Learning Systems in Dynamic Environments} 
\author{
	Kaan~Gokcesu, Hakan~Gokcesu
}
\maketitle

\begin{abstract}
	Sequential learning systems are used in a wide variety of problems from decision making to optimization, where they provide a 'belief' (opinion) to nature, and then update this belief based on the feedback (result) to minimize (or maximize) some cost or loss (conversely, utility or gain). The goal is to reach an objective by exploiting the temporal relation inherent to the nature's feedback (state). By exploiting this relation, specific learning systems can be designed that perform asymptotically optimal for various applications. However, if the framework of the problem is not stationary, i.e., the nature's state sometimes changes arbitrarily, the past cumulative belief revision done by the system may become useless and the system may fail if it lacks adaptivity. While this adaptivity can be directly implemented in specific cases (e.g., convex optimization), it is mostly not straightforward for general learning tasks. To this end, we propose an efficient optimal mixture framework for general sequential learning systems, which we call the recursive experts for dynamic environments. For this purpose, we design hyper-experts that incorporate the learning systems at our disposal and recursively merge in a specific way to achieve minimax optimal regret bounds up to constant factors. The multiplicative increases in computational complexity from the initial system to our adaptive system are only logarithmic-in-time factors.  
\end{abstract}

\section{Introduction}\label{sec:intro}
\subsection{Motivation}
The sequential learning systems \cite{cesa_book,poor_book} are central in a wide variety of fields from decision theory \cite{tnnls4}, game theory \cite{tnnls1,chang} and control theory \cite{tnnls3} to signal processing \cite{ozkan}, multi-agent systems \cite{vanli} and optimization \cite{zinkevich,hazan}. They are heavily used in various kinds of applications such as density estimation and source coding \cite{gDensity,willems,coding1,gAnomaly,coding2}, anomaly and outlier detection \cite{gokcesu_anomaly2,gIncremental}, adversarial bandits \cite{cesa-bianchi,cesa2007,gBandit} and prediction \cite{singer,singer2,gokcesu_prediction}. 
 
In general, the learning systems (learners, algorithms) provide a 'belief' (an estimation, opinion, advice or a decision) that acts upon the nature; and based on the feedback (result) received, they update their beliefs to minimize (maximize) some cost or loss (utility or gain). Hence, a sequential learning system produces a belief $\xt$ at time $t$ and based on an observation, which measures how well (or accurate) this belief is, it updates its belief, i.e., the next belief $\xt[t+1]$ is created with the past beliefs and the corresponding incurred losses. The sequential learning system needs to utilize temporal reasoning since it may work on different tasks throughout the time horizon and the loss functions may be different at each time instance $t$. Therefore, to model the time dependency, the incurred loss at time $t$ is more correctly represented as $\lt(\xt)$. The goal is to reach the objective (e.g., minimize the loss) by exploiting the inherent temporal relation \cite{cesa_book}.
To this end, for various applications, specific learning systems can be designed that perform asymptotically optimal that can achieve the 'best' belief in hindsight. As an example, suppose that we are trying to infer (learn) the mean $\mu^*$ of a Gaussian process with known variance $\sigma_0$. We have a sequential learning system that produces, at time $t$, the belief $\mu_t$, which is the sample mean of the observations up to $t$. Then, asymptotically, this system will accurately model the mean since its estimations $\mu_t$ converge to the true mean $\mu^*$ by the law of large numbers \cite{grimmett}. 

On the other hand, if there exist (possibly unplanned) changes in the nature (e.g., its state changes arbitrarily), the past cumulative belief revision done by the system may become useless and the system may fail if it lacks adaptivity. While the system utilizes the temporal relation in its belief updates throughout the time horizon, this behavior may even have adversarial effects in the system if the temporal relation no longer holds. As an example, the previously mentioned learning system will fail to accurately estimate the mean if it changes arbitrarily throughout the time horizon. Certain precautions can be taken to prevent this situation (e.g., in this example, we can instead take a weighted sample mean to produce our belief where we give more emphasis to the more recent observations). This adaptivity can be directly implemented in some specific cases. As an example, in the online convex optimization problem, tuning some parameter (e.g., the learning rate $\eta$) of the learning system (optimizer) can make it more adaptive \cite{zinkevich}. However, it is not straightforward for general learning tasks, and learning in an environment with changes in the nature may prove to be fundamentally different from learning in an environment with no such change \cite{cesa_book, poor_book}. 

\subsection{Learning Environments}
To better understand this, we need to analyze and distinguish the different learning environments that we may come across in the literature; which are static, drifting and dynamic.

\subsubsection{Static Environment}
Static environment is self-explanatory in that the environment does not change and the goal is to learn a specific optimal parameter $\xt[*]$; such as the aforementioned example of learning the mean $\mu^*$ of a Gaussian process. The research on this alone is abundant and its applications are numerous \cite{cesa_book}. In general, the performance measure is done by the notion of 'regret', which is defined as the difference between the loss of the learner and loss of the 'best' parameter selected in hindsight \cite{vovk,hazan}. 

\subsubsection{Drifting Environment}
Drifting environment, on the other hand, differs from the static environment such that the environment is changing slowly over time. A follow-up example would be the tracking of the mean $\mu_{t}^*$ of a Gaussian process that continuously (albeit slowly) changes over time. There are varies research on this topic as well because of its interesting applications \cite{hall2013, jadbabaie2015, rakhlin2013}, which are generally focused on the online convex optimization \cite{zinkevich,gokcesu2019efficient}. 
\subsubsection{Dynamic Environment}
Dynamic environment differs from them both such that the change in nature is allowed to be arbitrary as opposed to the slow or no change in the drifting and the static environments, respectively. A follow-up example would be the estimation of the mean $\mu^*_i$ of a Gaussian process that may arbitrarily change throughout time. When the environment undergoes many changes, the previous notion of regret may cease to be the best measure of performance. Since the regret-minimizing algorithms originating from the static environments become stagnant in such scenarios, the concept of regret needed to be generalized to allow for changing prediction strategies \cite{comp2,freund1997,lehrer2003,blum2007,hazan2009}. 

The static environment can be considered as a special case of the dynamic environment itself. Moreover, although the drifting scenario is an interesting setting in the convex optimization problem, it may not be meaningful in a general learning task. Some problems may be highly sensitive to the parameter we learn and even a small change may require the learning to be done almost from scratch. Henceforth, our focus will be on the dynamic environments for a more general analysis and algorithm development, since both the static and the drifting environments can be considered branches of the more general dynamic environment. 

\subsection{Going From Static to Dynamic}
Our goal in this work is to incorporate adaptivity into varying applications and problems that we are able to provide static solutions for. However, in general, having a solution to the static version of a problem does not always imply a solution to its dynamic counterpart. Even though we may have a suitable approach to solve a given problem in a static environment, we may not be able to incorporate that for our use in a dynamic environment in a straightforward manner \cite{cesa_book,poor_book}. 

For this purpose, there are two seemingly distinct approaches in literature, which are the sleeping (specialist) experts and the restarting experts.

\subsubsection{Sleeping Experts}
The sleeping experts \cite{chernov2009} work by creating a pool of
virtual experts, where for each real expert $n$, a virtual expert is included
that mimics the learner's behavior in the first $t-1$ trials (which is another way
to say that this expert is a specialist \cite{freund1997using} that abstains from prediction, or sleeps, during the first $t-1$ trials), and predicts as expert $n$ from trial $t$ onward. Then, these experts are mixed for adaptivity. These sleeping experts can be generalized to use different time selection functions as well \cite{blum2007}. 

\subsubsection{Restarting Experts}
The restarting experts \cite{hazan2009} works by starting with a base algorithm and restarting a copy of it each trial. Then, by aggregating the predictions of these copies adaptivity is achieved. 

Although conceptually dissimilar, it is shown that both of these approaches reduce to the same algorithm with variable parameters. Specifically, it has been shown that both of these constructions reduce to the Fixed Share algorithm \cite{adamskiy} with variable switching rates.

\subsection{General Techniques}
Nevertheless, it all comes down to the selection of a set of time intervals that run a suitable base algorithm and the aggregation of their beliefs. Thus, the algorithms have two parts, which are:
\subsubsection{Time Interval Selection}
Optimization of the time intervals is a bit tricky. On one hand, the set of interval should be large so that for every possible interval there exists an expert that works well. On the other hand, the number of intervals should be small, since running many experts in parallel will result in high computation cost. \cite{gyorgy2012} and \cite{zhang2018} have developed new ways to construct intervals which can trade effectiveness for efficiency explicitly. Especially, the exponential quantization of the intervals for efficient covering has been popular such as the variants of the geometric coverage (GC) \cite{daniely}, compact geometric coverage (CGC) \cite{zhang2019} and the dense geometric coverage (DGC) \cite{zhang2020}.
\subsubsection{Aggregation}
There are numerous techniques from multiplicative weights \cite{daniely} to exponential weights \cite{hazan2009,gokcesu2020generalized} for the aggregation, which is heavily investigated by itself \cite{littlestone1994}. 

\subsection{Organization}
Nonetheless, most of the research is focused on convex, exp-concave or smooth functions, and mention of the generalized learning problems/settings are minimal. To this end, we develop algorithms for efficient and optimal merger of general learning systems that incorporate adaptivity to compete in dynamic environments. We develop our algorithms in sequence to show the decisions made in each step of the process. We show that most of the work in literature coincide naturally with our development. We start by a direct aggregation of exponentially quantized time intervals. We follow that by modifying the aggregation to a switching mixture of experts. Finally, we propose an approach which we call the Recursive Experts where the complete mixture is inherent.
The organization of our paper is as follows. 

\subsubsection{Preliminaries}
In Section \ref{sec:prelim}, we provide some important preliminary information that covers the problem definitions, solution types and a general base algorithm. 

\subsubsection{Sub-Optimal Algorithm}
In Section \ref{sec:logsub}, we start by developing the most straightforward (simplest) method that can achieve adaptivity, which utilizes resetting the base algorithm and aggregating for different reset policies. This approach has a sub-optimal performance because of the mismatch between the reset times and the times the environment changes. 

\subsubsection{Near-Optimal Algorithm}
In Section \ref{sec:log2near}, we utilize the fact that the cumulative reset policies can be chosen to cover the whole time horizon and aggregate the different policies with a mixture of switching experts framework. This approach has a near-optimal performance (up to logarithmic factors). 

\subsubsection{Optimal Algorithm}
In Section \ref{sec:lognear}, we propose an algorithm that inherently optimizes its internal parameters to achieve better performance and efficiency. Instead of the standard mixture approaches, we design a recursive mixture strategy and succeed in creating an efficient algorithm with optimal performance.

\subsubsection{Conclusion}
We finish with some remarks in Section \ref{sec:conc}.

\section{Preliminaries}\label{sec:prelim}
\subsection{Static Problem}\label{sec:staticprob}
We start the problem definition by first explaining the 'static' problem. Let $\xt$ be the belief we produce at time $t$, and let
\begin{align}
	\xt[1]^T\triangleq\{\xt[1],\ldots,\xt[T]\}\label{x1T}
\end{align}
be our beliefs from $t=1$ to $T$. Let $\lt(\xt)$ be a finite loss we incur from our belief $\xt$, then, the cumulative loss in $T$ is
\begin{align}
L_T(\xt[1]^T)\triangleq &\sum_{t=1}^{T}\lt(\xt).\label{Lt}
\end{align} 
Let $x_*$ be the best 'static' belief chosen in hindsight, i.e.,
\begin{align}
	\xt[*]=\argmin_{x}\sum_{t=1}^{T}\lt(\xt[])\label{x*},
\end{align}
hence, its cumulative loss is given by
\begin{align}
L_T(\xt[*])\triangleq &\sum_{t=1}^{T}\lt(\xt[*]).\label{Lx*}
\end{align} 
Thus, the regret of this static problem is the difference of the cumulative losses in \eqref{Lt} and \eqref{Lx*}, which is
\begin{align}
	R(T,x^*)\triangleq L_T(\xt[1]^T)-L_T(\xt[*]).\label{RTx*}
\end{align} 

\subsection{Dynamic Problem}\label{sec:dynamicprob}
Next, we explain the dynamic version of the general belief selection problem. While our selections $\xt[1]^T$ in \eqref{x1T} and their cumulative loss $L_T(\xt[1]^T)$ in \eqref{Lt} remain the same, the competition differs. Instead of the best 'static' selection let us compete against the best 'dynamic' selection in hindsight. Let $C$ denote the number of times the underlying optimal parameter $\xt[*]$ changes throughout the time horizon. Let us define $\xt[*]^{(c)}$ for $c\in\{1,\ldots,C\}$ as the optimal selections in hindsight, which are individually optimal during $C$ distinct time segments that cover the time horizon $T$. Let $t_c$ denote the length of the $c^{th}$ segment and $T_c$ be the cumulative sum of $t_{c'}$ from $c'=1$ to $c$. Then, we have the following expression instead of \eqref{Lx*}
\begin{align}
L_T(\{x_*^{(c)},t_c\}_{c=1}^C)=\sum_{c=1}^C\sum_{T_{c-1}+1}^{T_c}l_t(x_*^{(c)}),
\end{align}
where $T_c$ is $0$ for $c=0$. Thus, our regret in \eqref{RTx*} becomes
\begin{align}
	R(T,\{x_*^{(c)},t_c\}_{c=1}^C)\triangleq L_T(\xt[1]^T)-L_T(\{x_*^{(c)},t_c\}_{c=1}^C).\label{RTx*c}
\end{align}

\subsection{Base Algorithm: A Learning System for the Static Problem}
\begin{algorithm}[!t]
	\caption{A General Learning System}\label{alg:belief}
	{\begin{algorithmic}[1]
			\STATE Initialize internal parameters, $\xt[1]$
			\FOR {$t=1,2,\ldots$}
			\STATE Observe $\lambda_t$, $\lt(\xt)$
			\STATE Determine $\ft[t](\cdot)$
			\STATE Update $\xt[t+1]=\ft(\xt)$
			\ENDFOR
	\end{algorithmic}}
\end{algorithm}
Let us have an algorithm (a general learning system) at hand, that can solve the problem in Section \ref{sec:staticprob} (hopefully optimally, i.e., provides an optimal order regret bound with an optimal order computational complexity), which works as the following. A general learning system (or learner) produces its belief $\xt$ at time $t$ and incurs the loss $\lt(\xt)$ at time $t$. The learning system updates its beliefs such that $\xt[t+1]$ is a function of its past beliefs, i.e., $\{\xt[t],\xt[t-1],\ldots,\xt[1]\}$, and losses, i.e., $\{\lt(\xt[t]),\lt[t-1](\xt[t-1]),\ldots,\lt[1](\xt[1])\}$ and some auxiliary information, i.e., $\{\lambda_t,\ldots,\lambda_1\}$. Thus,
\begin{align}
\xt[t+1]=f(\xt,\ldots,\xt[1];\lt[t](\xt),\ldots,\lt[1](\xt[1]);\lambda_t,\ldots,\lambda_1).\label{xt}
\end{align}
The update function $f(\cdot)$ is designed based on the underlying problem setting, and can differ between various applications. For sequential learning systems, the update is only dependent on the current belief, observation and the algorithm's current state. Hence, the update becomes
\begin{align}
\xt[t+1]=\ft[t](\xt),
\end{align}
for some $\ft(\cdot)$, which models the algorithm's state at time $t$.
As an example, the gradient-based sequential first-order methods are commonly used for convex optimization problems \cite{zinkevich}, where the update is given by 
\begin{align}
\xt[t+1]=\ft[t](\xt)=\xt-\eta_t\nabla_{\xt}\lt(\xt),
\end{align}  
where the gradient is part of the auxiliary information $\lambda_t$. 
A summary of the algorithm is in Alg. \ref{alg:belief}.

To investigate the performance of the algorithms, we need to study their regret bounds as in Section \ref{sec:staticprob} and \ref{sec:dynamicprob}. For the problem in Section \ref{sec:staticprob}, let our learning system in Alg. \ref{alg:belief} have the following regret bound.
\begin{assumption} \label{ass:RB}
	For the static problem in \eqref{RTx*}, let Alg. \ref{alg:belief} have the regret bound
	\begin{align}
	R_{A\ref{alg:belief}}(T)= O(T^{1-\alpha}),\nonumber
	\end{align}
	where $0<\alpha\leq0.5$ and $O(\cdot)$ is the big-O notation. Hence, $R_{A\ref{alg:belief}}(T)\leq KT^{1-\alpha}$ for some finite $K$ for all $T$.
\end{assumption}

\begin{remark}
	In a wide range of problems where adaptivity is straightforward (such as the convex optimization and linear games), we have $O(\sqrt{T})$ regret bounds \cite{zinkevich,cesa-bianchi,cesa_book}. Thus, as inferred from Assumption \ref{ass:RB}, we focus on the 'at least as hard' problems, where
	\begin{align}
		R_{A\ref{alg:belief}}(T)= \Omega(\sqrt{T}),\nonumber
	\end{align} 
	where $\Omega(\cdot)$ is the Big-Omega notation. 
\end{remark}

Although, this learning system is able to learn an optimum fixed parameter (the static problem), it may, in general, fail to learn an optimal parameter which is dynamically changing throughout the time horizon (the dynamic problem). For the problem in Section \ref{sec:dynamicprob}, we can only use Alg. \ref{alg:belief} by itself, if we know the time instances the optimal parameter changes (i.e., the exact location of the time segments where $\xt[*]^{(c)}$ are individually optimal for $c\in\{1,\ldots,C\}$) so that we can reset our learning system at these specific time instances.

\subsection{Best Performance Achievable with the Base Algorithm}\label{sec:baseC}
Suppose the regret of the base algorithm for a given static problem is as in Assumption \ref{ass:RB}. Then, for the dynamic problem, we have the following.
\begin{corollary}\label{cor:opt}
	We have the following optimal regret bound for the dynamic problem in \eqref{RTx*c} when using Alg. 1
	\begin{align}
	R_{A\ref{alg:belief}.*}(T,C)=O(C^{\alpha}T^{1-\alpha}).\nonumber
	\end{align} 
	\begin{proof}
		The dynamic version of the problem as given in Section \ref{sec:dynamicprob} that divides the time horizon in to $t_c$ length parts for $c\in\{1,\ldots,C\}$, will have the following optimum regret bound:
		\begin{align}
		R_{A\ref{alg:belief}.*}(T,\{x_*^{(c)},t_c\}_{c=1}^C)\triangleq\sum_{c=1}^{C}R_{A\ref{alg:belief}}(t_c).
		\end{align} 
		Since the bound in Assumption \ref{ass:RB} is concave, we have
		\begin{align}
		R_{A\ref{alg:belief}.*}(T,C)\leq KC\left(\frac{T}{C}\right)^{1-\alpha},
		\end{align} 
		for some finite $K$, then
		\begin{align}
		R_{A\ref{alg:belief}.*}(T,C)=O(C^{\alpha}T^{1-\alpha}),
		\end{align}
		which concludes the proof.
	\end{proof}
\end{corollary}

Thus, the result in Corollary \ref{cor:opt} is the optimal regret bound achievable by using Alg. \ref{alg:belief}, where the base algorithm may or may not be optimal itself. Note that, the number of changes $C$ should be sublinear for viable learning (sublinear regret).

\subsection{Algorithm Types}
Here, we categorize the algorithms according their performance optimality in accordance with performance of the base algorithm in Section \ref{sec:baseC}. The optimality types in accordance with the base algorithm regret are as follows:

\subsubsection{Optimal Algorithms}
If an algorithm has the same order of bounds as in Corollary \ref{cor:opt}, we call it an optimal algorithm.
\begin{definition}\label{def:RAopt}
	An optimal algorithm has the regret bound
	\begin{align}
	R_{AO}(T,C)=O\left(C^{\alpha}T^{1-\alpha}\right).\nonumber
	\end{align}
\end{definition}

\subsubsection{Near-optimal Algorithms}
If an algorithm achieves the same order of regret bounds as the result in Corollary \ref{cor:opt} up to logarithmic terms, we call it a near-optimal algorithm.
\begin{definition}\label{def:RAnear}
A near-optimal algorithm has the regret bound
\begin{align}
R_{AN}(T,C)=\tO\left(C^{\alpha}T^{1-\alpha}\right),\nonumber
\end{align}
where $\tO(n)$ is the soft-O notation, which is a shorthand for $O(n \log^k(n)) \text{ for some finite } k$,
i.e., soft-O notation ignores the logarithmic terms.
\end{definition}

\subsubsection{Sub-optimal Algorithms}
Even if we do not have the regret bounds in Definition \ref{def:RAopt} or \ref{def:RAnear}, our algorithm should still have sublinear regret bounds, i.e., $o(T)$, where $o(\cdot)$ is the Little-O notation. 
When we have such a regret bound, we have Hannan consistency \cite{hart},
which is always needed to say we are able to solve the problem (provide a viable solution). 
\begin{definition}
	A sub-optimal algorithm has the regret bound
	\begin{align}
		R_{AS}(T,C)=o(T),\nonumber
	\end{align}
	which is not optimal but is still sublinear.
\end{definition}

\subsection{Expected Regret vs High Probability Regret}
In the remainder of the paper, we will focus on the expected regret bounds, which may also be referred as simply by regret bounds. The high probability bounds are straightforward for any mixture algorithm when the losses are bounded (as we assume) because of the following Lemma. 

\begin{lemma}
	For any algorithm with the expected regret $R_A(T)$, we have with probability $1-\delta$, the following regret bound
	\begin{align}
		R_{A.1}(T,\delta)=O\left(\sqrt{\log(\delta^{-1})T}\right)+R_A(T),
	\end{align}
	for $0<\delta<1$ if the losses are bounded.
	\begin{proof}
		The proof comes from the Hoeffding's Lemma \cite{hoeffding1994}.
	\end{proof}
\end{lemma}

\subsection{Unknown Time Horizon $T$}
In the remainder of the paper, we will study the regret bounds of an algorithm for known time horizon $T$, since, for unknown time horizon, we can utilize the doubling trick and run that algorithm for the time lengths that are powers of $2$ in succession. The bounds on unknown $T$ is similar because of the following Lemma.
\begin{lemma}\label{lem:unkT}
	For any algorithm with the expected regret bound $$R_{AC}(T,C)=R_0(T,C)C^\alpha T^{1-\alpha},$$ where $R_0(T,C)$ is nondecreasing in the number of changes $C$ and the known time horizon $T$, we have the following regret bound for an unknown time horizon $T$.
	\begin{align}
	R_{AC.1}(T,C)=O\left(R_{AC}(T,C)\right).\nonumber
	\end{align}
	\begin{proof}
		Let us run that algorithm for time horizon $T_i=2^i$ for $i\in\{0,1,\ldots\}$. Let $C_i$ be the number of changes in the time horizon $T_i$. Let us stop at an arbitrary time $T$ such that $T_I$ is the last time horizon used for the algorithm. The cumulative regret will be bounded by
		\begin{align}
		R_{AC.1}(T,C)\leq&R_{AC.1}(2T_I-1,C)\\
		\leq&\sum_{i=0}^{I}R_{AC}(T_i,C_i),
		\end{align} 
		since the regret bound is monotonic. From $T_I\leq T\leq 2T_I-1$, $C-1\geq\sum_{i=0}^{I}(C_i-1)$ and the definition of $R_{AC}(\cdot,\cdot)$, we have
		\begin{align}
		R_{AC.1}(T,C)\leq\left(\max_{0\leq i\leq I}R_0(T_i,C_i)\right)\sum_{i=0}^IC_i^{\alpha}T_i^{1-\alpha}
		\end{align} 
		From monotonicity of $R_0(\cdot,\cdot)$ and the fact that $0<\alpha\leq0.5$, we have
		\begin{align}
		R_{AC.1}(T,C)\leq R_0(T,C)\sum_{i=0}^I\left((C_i-1)^{\alpha}T_i^{1-\alpha}+T_i^{1-\alpha}\right)\label{eq:6}
		\end{align} 
		Since the maximizing $(C_i-1)$'s in \eqref{eq:6} are proportionate to $T_i$, and $T_i^{1-\alpha}$ is a power series, we end up with
		\begin{align}
		R_{AC.1}(T,C)\leq O\left(R_0(T,C)\left(C^\alpha T^{1-\alpha}+T^{1-\alpha}\right)\right),
		\end{align}
		which concludes the proof since $C\geq 1$.
	\end{proof}
\end{lemma}

\section{A Log-complexity Sub-optimal Algorithm}\label{sec:logsub}
In this section, we start by providing a sub-optimal algorithm whose computational complexity has $\log(T)$ overhead for the time horizon $T$. To this end, we start by making the algorithm in Alg. \ref{alg:belief} more adaptive.
\subsection{Resetting the Algorithm for Adaptivity}
\begin{algorithm}[!t]
	\caption{A General Learning System with Reset}\label{alg:reset}
	{\begin{algorithmic}[1]
			\STATE Set $t_r$, i.e., the reset period
			\STATE Start Alg. \ref{alg:belief}
			\STATE Set $x_t$ as the belief of Alg. \ref{alg:belief}
			\FOR {$t=1,2,\ldots,T$}
			\STATE Observe $\lt(\xt)$
			\IF{$t_r$ divides $t$}
			\STATE Restart, Alg. \ref{alg:belief}
			\STATE Reset $x_t$ as the belief of Alg. \ref{alg:belief}
			\ENDIF
			\ENDFOR
	\end{algorithmic}}
\end{algorithm}
A straightforward approach to incorporate adaptivity into the general learning system in Alg. \ref{alg:belief} is to utilize the resetting behavior, i.e., forcing the algorithm to reset so that it reinitializes its internal parameters to forget the unrelated (or even adverse) information that is potentially acquired by its past learning. 

Suppose the optimal performance can be achieved by resetting the algorithm in $C$ number of time instances such that they divide the time horizon $T$ into segments of length $t_c$ for $c\in\{1,\ldots,C\}$ as in the dynamic problem in Section \ref{sec:dynamicprob}. Even though it is possible to acquire the optimal performance by resetting the algorithm at time instances $T_c$ (as stated in Section \ref{sec:baseC}), in general, we do not have access to $T_c$. Henceforth, to start up, we shall naturally investigate the possible performance when we make the algorithm periodically reset, i.e., reset every $t_r$ times. We can also consider different resetting policies, however, periodic reset makes most sense since we have no knowledge of the actual reset times whatsoever. Moreover, this also coincides with the time interval selection in \cite{gyorgy2012,daniely,zhang2018,zhang2019,zhang2020}. A summary of the algorithm is given in Alg. \ref{alg:reset}. 

We point out that selecting a suitable reset period is not trivial, where the possible integer reset periods we can select during a time horizon $T$ are $t_r\in\{2,\ldots,T\}$ (the trivial reset time $t_r=1$ is useless since it does not ever make a belief update, i.e., learn, and will always produce the initial belief $\xt[1]$ in Alg. \ref{alg:belief}). The question is which reset period would perform best for our purposes. While choosing the reset period too long may result in insufficient adaptivity and choosing it too short result in insufficient learning.
If we have access to the number of changes $C$ or at least an upper bound $C_u$, we may optimize $t_r$ accordingly. However, we may not even have an access to a nontrivial $C_u$; and, even though we do have access to $C_u$, this may substantially differ from the true $C$. Thus, to this end, we incorporate a belief merging scheme.

\subsection{Parallel Belief Merging for Optimization}\label{sec:parallel}
\begin{algorithm}[!t]
	\caption{Parallel Belief Merging}\label{alg:parallel}
	{\begin{algorithmic}[1]
			\STATE Inputs $T$
			\STATE Set $N=\ceil{\log_2 T}$
			\STATE Set $\eta=\sqrt{\frac{\log N}{T}}$
			\STATE Start $N$ copies of Alg. \ref{alg:reset} with $t_r=2^i$, $i\in\{1,\ldots,N\}$ 
			\STATE Initialize $\pt[i,1]=\frac{1}{N}$
			\STATE Set $\xt[i,t]$ as the belief of $i^{th}$ Alg. \ref{alg:reset} at time $t$
			\FOR {$t=1,2,\ldots,T$}
			\STATE Draw $I$ from the distribution $\pt[i,t]$
			\STATE Set $\xt=\xt[I,t]$
			\STATE Observe $\lt(\xt[i,t])$ for $i\in\{1,\ldots,N\}$ 
			\STATE $\tpt[i,t+1]=\pt[i,t]e^{-\eta l_t(\xt[i,t])}$
			\STATE $	\pt[i,t+1]=\dfrac{\tpt[i,t+1]}{\sum_{j=1}^N\tpt[j,t+1]}$
			\ENDFOR
	\end{algorithmic}}
\end{algorithm}
We consider all possible reset periods and run in parallel different versions of the learning system each of which utilizes a different reset period. We then combine their beliefs in a mixture of experts framework \cite{doubling_trick} to achieve the performance of the best reset period, which coincides with the approach of a direct aggregation of the hyper-experts. However, running in parallel $T$ copies (number of all possible reset periods) of the learning system will increase the computational complexity by linear-in-time factors. Therefore, instead of mixing all possible reset periods, we naturally mix only the reset periods that are powers of $2$. If we only mix the reset times from this specific set, we will only increase the computational complexity with a logarithmic-in-time factor, which coincides with the exponential quantization of the time intervals as in the geometric coverage (GC) \cite{daniely}, compact geometric coverage (CGC) \cite{zhang2019} and the dense geometric coverage (DGC) \cite{zhang2020}. We force the $i^{th}$ algorithm, where $i\in\{1,\ldots,N\}$ and $N=\ceil{\log_2T}$, to make a reset every $2^i$ time. 

Let $\xt[i,t]$ denote the belief of the $i^{th}$ parallel running algorithm at time $t$.
Let us have the probability simplex $\pt[i,t]$, where $\sum_{i=1}^{N}\pt[i,t]=1$ and $N=\ceil{\log_2T}$.
We create our belief $\xt$ by randomly selecting $\xt[i,t]$ with the probabilities $\pt[i,t]$. Thus, our expected belief and expected loss are
\begin{align} \label{eq:Ex}
\expect{\xt}=\sum_{i=1}^N\pt[i,t]\xt[i,t], &&\expect{\lt(\xt)}=\sum_{i=1}^N\pt[i,t]\lt(\xt[i,t]).
\end{align}
We create the probabilities $\pt[i,t]$ with the exponentiated losses as in the exponential weights algorithm \cite{hazan2009}. Thus, $\pt[i,t]$ for $i\in\{1,\ldots,N\}$ are updated as
\begin{align}
\tpt[i,t+1]=\pt[i,t]e^{-\eta l_t(\xt[i,t])},\label{pt}
\end{align}       
where $\eta$ is the learning rate.
New probabilities are given by the normalization of \eqref{pt} as
\begin{align}
\pt[i,t+1]=\dfrac{\tpt[i,t+1]}{\sum_{j=1}^N\tpt[j,t+1]}.
\end{align}
A summary of the algorithm is given in Alg. \ref{alg:parallel}. We next study its regret bounds.

\subsection{Performance Analysis}
For performance analysis, we first study the regret bound of Alg. \ref{alg:reset} with an optimized $t_r$.

\begin{lemma}\label{lem:A1C}
	Alg. \ref{alg:reset} has the following regret bound when its parameter $t_r$ is optimized with the number of changes $C$,
	\begin{align}
	R_{A\ref{alg:reset}}(T,C)=&O(T^{\frac{1}{1+\alpha}}{C}^{\frac{\alpha}{1+\alpha}}),\nonumber
	\end{align}
	where $\alpha$ is as in Assumption \ref{ass:RB}.
	\begin{proof}
		Suppose our algorithm resets every $t_r$ rounds. Then, our cumulative regret will be generalized by the following.
		\begin{align}
		R_{A\ref{alg:reset}}(T,C)=O\left(\frac{T}{t_r}t_r^{1-\alpha}\right)+O\left(Ct_r\right),\label{RBC}
		\end{align}
		where the first part of \eqref{RBC} is the sum of the regret bound for $T/t_r$ runs of Alg. \ref{alg:belief}, and the second part results from the mismatch between actual reset time and the optimal reset times assuming the loss is bounded. To minimize \eqref{RBC}, we set
		\begin{align}
		t_r^*\sim\left(\frac{\alpha T}{C}\right)^{\frac{1}{1+\alpha}},
		\end{align}
		which gives
		\begin{align}
		R_{A\ref{alg:reset}}(T,C)=&O(T^{\frac{1}{1+\alpha}}{C}^{\frac{\alpha}{1+\alpha}})\label{RBr},
		\end{align}
		and concludes the proof.
	\end{proof}
\end{lemma}
The result in Lemma \ref{lem:A1C} is Hannan consistent (i.e., sublinear) as long as $C$ is sublinear, which gives us an able albeit sub-optimal algorithm. However, we still need the number of changes $C$ to optimize $t_r$. If we do not have access to $C$, but an upper bound $C_u$, we can instead use it to optimize $t_r$.
\begin{corollary}\label{cor:A1Cu}
	Alg. \ref{alg:reset} has the following regret bound when $C_u\geq C$ is used to optimize $t_r$
	\begin{align}
	R_{A\ref{alg:reset}}(T,C)=&O(T^{\frac{1}{1+\alpha}}{C_u}^{\frac{\alpha}{1+\alpha}})\label{RBru}.
	\end{align}
	\begin{proof}
		The result follows from Lemma \ref{lem:A1C} since $C_u\geq C$.
	\end{proof}
\end{corollary}
By utilizing the result of Corollary \ref{cor:A1Cu}, we get the following Theorem, which provides a bound for unknown $C$. 
\begin{theorem}\label{thm:A3}
	Alg. \ref{alg:parallel} has the following regret bound
	\begin{align}
		R_{A\ref{alg:parallel}}(T,C)=O(T^{\frac{1}{1+\alpha}}{C}^{\frac{\alpha}{1+\alpha}})
	\end{align}
	\begin{proof}
		The performance of our belief merging is as follows
		\begin{align}
		R_{A\ref{alg:parallel}}(T)=O(\sqrt{T\log\log T})+O(T^{\frac{1}{1+\alpha}}{C}^{\frac{\alpha}{1+\alpha}}).\label{R1}
		\end{align}
		where the first term in \eqref{R1} is the mixture redundancy when the learning rate $\eta$ is set as the following \cite{doubling_trick}
		\begin{align}
		\eta^*\sim\sqrt{\frac{\log\log T}{T}},
		\end{align}
		and the second term in \eqref{R1} is the result from Corollary \ref{cor:A1Cu} since in the mixture there is a $C_i$ such that $C\leq C_i\leq 2C$. Since $\alpha\leq0.5$ and $\log\log(T)$ is sub-polynomial, we have
		\begin{align}
		R_{A\ref{alg:parallel}}(T)=O(T^{\frac{1}{1+\alpha}}{C}^{\frac{\alpha}{1+\alpha}}),
		\end{align}
		which concludes the proof.
	\end{proof}
\end{theorem}
Thus, the result in Lemma \ref{lem:A1C} is achieved without knowing $C$ at all, and we have created a sub-optimal algorithm (i.e., $o(T)$ regret) with $\log(T)$ computational complexity overhead.

\section{A $\text{Log}^2$-complexity Near-optimal Algorithm}\label{sec:log2near}
We observe that the reason for the suboptimal regret in Section \ref{sec:logsub} is rooted in the mismatch between the optimal parameters change times $T_c$ as defined in Section \ref{sec:dynamicprob} and the algorithm reset times, which are periodical. To solve this with the parallel aggregation, we need to consider all possible algorithm reset times and mix them together. However, this approach would substantially increase the computational complexity. Instead, we observe that because of the exponential quantization of reset times, all possible $T_c$ instances are already covered \cite{daniely}. We utilize this to obtain better bounds.

\subsection{First-Level: Shared Merging for Adaptivity}
\begin{algorithm}[!t]
	\caption{First-Level: Shared Merging for Adaptivity}\label{alg:firstlevel}
	{\begin{algorithmic}[1]
			\STATE Inputs $C$, $T$
			\STATE Set $\sigma=CT^{-1}$
			\STATE Set $\eta=\sqrt{CT^{-1}\log(T/C)}$
			\STATE Set $N=\ceil{\log_2T}$
			\STATE Start $N$ copies of Alg. \ref{alg:reset}, each indexed by $i\in\{1,\ldots,N\}$ with reset period $2^i$ and $1^{st}$ reset time $2^{i-1}$
			\STATE Initialize $\pt[1,1]=1$, $\pt[i,1]=0$ for $i\in\{2,\ldots,N\}$
			\STATE Set $\xt[i,t]$ as the belief of $i^{th}$ Alg. \ref{alg:belief} at time $t$
			\FOR {$t=1,2,\ldots,T$}
			\STATE Draw $I$ from the distribution $\pt[i,t]$
			\STATE Set $\xxt=\xt[I,t]$
			\STATE Observe $\lt(\xt[i,t])$ for $i\in\{1,\ldots,N\}$
			\STATE Find $j$ such that $2^{j}k+2^{j-1}=t+1$ for some $k$ 
			\STATE $\gt{j,t}=
			\sum_{i=1}^N\pt[i,t]e^{-\eta l_t(\xt[i,t])}$, and $\gt{i,t}=0$ for $i\neq j$
			\STATE $\tpt[i,t+1]=(1-\sigma)\pt[i,t]e^{-\eta l_t(\xt[i,t])}+\sigma\gt{i,t}$
			\STATE $	\pt[i,t+1]=\dfrac{\tpt[i,t+1]}{\sum_{j=1}^N\tpt[j,t+1]}$
			\ENDFOR
	\end{algorithmic}}
\end{algorithm}
For this purpose, we make a few changes in Alg. \ref{alg:parallel}, to realize its full potential. We start the design similarly as in Section \ref{sec:parallel} by mixing the reset periods which are powers of $2$, where we force the $i^{th}$ algorithm to make a reset every $2^i$ time but with the first reset at time $2^{i-1}$. We again denote $\xt[i,t]$ as the belief of the $i^{th}$ algorithm at time $t$, and $\pt[i,t]$ as its selection probability, and $\xxt$ as our belief by randomly selecting $\xt[i,t]$ with $\pt[i,t]$.
We again create $\pt[i,t]$ with the exponentiated losses. Moreover, different from Alg. \ref{alg:parallel}, we also share the probabilities with each other to allow switches between the parallel running algorithms as in a switching mixture of experts framework \cite{signal1,signal2,kozat1,kozat2,gBandit}. This approach coincides with the weight updates of restarting and sleeping experts \cite{chernov2009,freund1997using,blum2007,hazan2009,adamskiy}. The update is
\begin{align}
\tpt[i,t+1]=(1-\sigma)\pt[i,t]e^{-\eta l_t(\xt[i,t])}+\sigma\gt{i,t},\label{tpt}
\end{align}       
where $\sigma$ is the probability sharing parameter, $\eta$ is the learning rate and $\gt{i,t}$ is the mixture weight, which is given by
\begin{align}
\gt{i,t}=
\begin{cases}
\sum_{i=1}^N\pt[i,t]e^{-\eta l_t(\xt[i,t])}, & \text{if $i$ resets at $t+1$}\\
0, & \text{otherwise}
\end{cases}.
\end{align}
New probabilities are similarly given by normalization. A summary of the algorithm is provided in Alg. \ref{alg:firstlevel}.

\subsection{Second-Level: Parallel Merging for Optimization}
\begin{algorithm}[!t]
	\caption{Second-Level: Parallel Merging for Optimization}\label{alg:secondlevel}
	{\begin{algorithmic}[1]
			\STATE Inputs $T,\eta$
			\STATE $N=\ceil{\log_2 T}$
			\STATE Start $N$ copies of Alg. \ref{alg:firstlevel}, indexed by $i\in\{1,\ldots,N\}$ with parameters $C=2^i$, $T=T$.
			\STATE Initialize $\ppt[i,1]=1/N$ for $i\in\{1,\ldots,N\}$
			\STATE Set $\xxt[i,t]$ as the belief of the $i^{th}$ Alg. \ref{alg:firstlevel}
			\FOR {$t=1,2,\ldots,T$}
			\STATE Draw $I$ from the distribution $\ppt[i,t]$
			\STATE $\xxxt=\xxt[I,t]$
			\STATE Observe $\expect{\lt(\xt[i,t]^{(1)})}$ for $i\in\{1,\ldots,N\}$
			\STATE Incur loss $\lt(\xxxt)$
			\STATE $\tppt[i,t+1]=\ppt[i,t]e^{-\eta' \expect{l_t(\xxt[i,t])}}$
			\STATE $\ppt[i,t+1]=\dfrac{\tppt[i,t+1]}{\sum_{j=1}^N\tppt[j,t+1]}.$
			\ENDFOR
	\end{algorithmic}}
\end{algorithm}
A straightforward setting of $\sigma$ and $\eta$ in Alg. \ref{alg:firstlevel} are \cite{cesa_book}
\begin{align}
\sigma_0=\dfrac{1}{T}, && \eta_0=\dfrac{1}{\sqrt{T}}.
\end{align}
However, if we know the number of switches we need to make between our $N$ parallel running algorithms (to account for the change in nature), then, we can optimize $\sigma$ and $\eta$.
\newcommand{\wC}{\widetilde{C}}

Let $\wC$ be the number of switches needed to be made between our algorithms for optimal performance. Then the optimal selection of the parameters $\sigma$ and $\eta$ are given by \cite{cesa_book, gBandit}
\begin{align}
\sigma_{\wC}=\dfrac{\wC}{T}, && \eta_{\wC}\sim\sqrt{\dfrac{\wC\log(T/\wC)}{T}}.\label{optPar}
\end{align}
However, one cannot know the number of switches $\wC$ needed to be made between our algorithms for optimal performance since the problem is sequential. Hence, to avoid the need for this a priori information, we consider the set of exponentially quantized number of switches $\wC\in\{2^0,2^1,\ldots\}$ and run copies of our first-level merging algorithm in Alg. \ref{alg:firstlevel} whose parameters are optimized with a particular value of $\wC$ and merge their beliefs similar in spirit to Alg. \ref{alg:parallel}. This approach increases the computational complexity by only $\log(T)$.

Let $\xxt[i,t]$ denote the belief of the $i^{th}$ parallel running first-level mixture at time $t$. Let us have the probability simplex $\ppt[i,t]$, where again $\sum_{i=1}^{N}\ppt[i,t]=1$ and $N=\ceil{\log_2T}$ (which is the number of first-level mixture algorithms run in parallel).
We create our belief $\xxxt$ by randomly selecting $\xxt[i,t]$ with the probabilities $\ppt[i,t]$. 
We create the probabilities $\ppt[i,t]$ with the exponentiated expected losses. Thus,
\begin{align}
\tppt[i,t+1]=\ppt[i,t]e^{-\eta' \expect{l_t(\xxt[i,t])}},\label{tppt}
\end{align} 
where $\eta'$ is the learning rate of the second-level mixture and $\expect{l_t(\xxt[i,t])}$ is the expected loss of the $i^{th}$ first-level mixture.
New probabilities are given by normalization and the summary of the algorithm is provided in Alg. \ref{alg:secondlevel}.

\subsection{Performance Analysis}
In this section, we will analyze the performance of our algorithm specifically in terms of the expected regret it incurs. We will build up the total expected regret our double mixture algorithm incurs, starting with the regret incurred from the second-level merging, then first-level merging, and finally from the learning systems with the reset periods.

\subsubsection{Regret of the Second-Level}
The regret of Alg. \ref{alg:secondlevel} derives similarly to Alg. \ref{alg:parallel} as shown in the following lemma.
\begin{lemma}\label{lem:A5}
	When the learning rate of Alg. \ref{alg:secondlevel} is set to
	\begin{align}
	\eta'=\sqrt{\dfrac{\log\log T}{T}},
	\end{align}
	we incur the following regret
	\begin{align}
	R_{A\ref{alg:secondlevel}}(T,C)\leq O\left(\sqrt{T\log\log T}\right)+R_{A\ref{alg:firstlevel}}(T,C),
	\end{align}
	where $R_{A\ref{alg:firstlevel}}(T,C)$ is the regret of the any first-level algorithm (i.e., Alg. \ref{alg:firstlevel}) in the mixture.
	\begin{proof}
		The proof follows from the proof of Theorem \ref{thm:A3}.
	\end{proof}
\end{lemma}
With a regret redundancy of $O\left(\sqrt{T\log\log T}\right)$, we achieve the performance of the best first-level algorithm in our mixture.

\subsubsection{Regret of the First-Level}
Let $\wC$ be the number of switches needed to be made between our parallel run of Alg. \ref{alg:reset} in Alg. \ref{alg:firstlevel} to achieve the minimum cumulative loss (i.e., the best switching path). 
We have the following regret $R_{A\ref{alg:firstlevel}}(T,C)$ incurred by the first-level mixture against the best switching path between the parallel-run learning systems \cite{comp2}.
\begin{lemma}\label{lem:A4}
	When Alg. \ref{alg:firstlevel} is run with some $C_i$ such that $\wC\leq C_i\leq 2\wC$, it has the following regret
	\begin{align} 
	{R_{A\ref{alg:firstlevel}}(T,C)}= O\left(\sqrt{T\wC\log (T/wC)}\right)+R_P(T,\wC),
	\end{align}
	where $R_P(T,C)$ is the regret of the best switching path. 
	\begin{proof}
		The proof follows from the regret of the switching experts \cite{gBandit,comp2}, where when the parameters are optimized as in \eqref{optPar}, we have the following
		\begin{align}
		{R_{A\ref{alg:firstlevel}}(T,C)}= O\left(\sqrt{T\wC\log (T/\wC)}\right)+R_P(T,\wC),
		\end{align}
		where $R_P(T,C)$ is the regret of the best switching path, which concludes the proof.
	\end{proof}
\end{lemma}

\subsubsection{Regret of the Learning Systems}
Let us denote the length of the $\tilde{c}^{th}$ time segment, which constitutes a single period of one of the parallel running Alg. \ref{alg:reset} in the mixture of Alg. \ref{alg:firstlevel} as $t_{\tilde{c}}\in\{2^0,2^1,2^2,\ldots\}$, where $\tilde{c}\in\{1,2,\ldots,\wC\}$. (note that, $\sum_{\tilde{c}=1}^{\wC}t_{\tilde{c}}=T$). For the cumulative regret resulting form Alg. \ref{alg:reset} in $\wC$ time segments, we have the following lemma.

\begin{lemma} \label{lem:P}
	The regret of the best switching path $R_P(T,\wC)$ of Alg. \ref{alg:reset} in Alg. \ref{alg:firstlevel} is
	\begin{align}
	R_P(T,\wC)\leq O\left({\wC}^{\alpha}T^{1-\alpha}\right),
	\end{align}
	\begin{proof}
		The regret of the best switching path $R_P(T,\wC)$ is given by the sum of the regrets from all the $\tilde{c}^{th}$ segments, which is
		\begin{align}
			R_P(T,\wC)\triangleq \sum_{\tilde{c}=1}^{\wC}R_{A1}(t_{\tilde{c}})
			\leq& \sum_{\tilde{c}=1}^{\wC}O(t_{\tilde{c}}^{1-\alpha})\\
			\leq& O\left(\wC^\alpha T^{1-\alpha}\right),
		\end{align}
		where we used Assumption \ref{ass:RB} and the concavity of $R_{A1}(\cdot)$ to conclude the proof.
	\end{proof}
\end{lemma}

\subsubsection{Total Regret of the Approach}
When a change in the optimal parameter happens, the best possible switching path in the first-level mixture will switch to the algorithm with the longest reset period that makes a reset at that time and continue switching to another algorithm with longer reset period whenever possible to minimize the regret. 
In the worst case scenario, we will have to first switch to the algorithm with the shortest reset time and build our way up from there, e.g., during the $c^{th}$ time segment, we will follow the following path: spend $1$ time in the first algorithm, then $2$ times in the second algorithm, then $4$ times in the third and so on. Hence, we have the following lemma.
\begin{lemma} \label{lem:wC}
	\begin{align}
	\wC\leq O(C\log(T/C))
	\end{align}
	\begin{proof}
		Let $n_c$ be the number of changes in the $c^{th}$ segment with length $t_c$, where the optimum parameter $x_*^{(c)}$ stay the same. Since $n_c\leq O(\log (t_c))$, we have
		\begin{align}
			\wC\triangleq \sum_{c=1}^C n_c
			\leq& \sum_{c=1}^C O(\log(t_c))\\
			\leq& O(C\log(T/C)),
		\end{align}
		from the concavity of logarithm, which concludes the proof.
	\end{proof}
\end{lemma}

To get the final results, all that is left is to combine Lemma \ref{lem:A5}, \ref{lem:A4}, \ref{lem:P} and \ref{lem:wC}.

\begin{theorem}\label{thm:A5}
	Alg. \ref{alg:secondlevel} has the following regret bound
	\begin{align}
		R_{A\ref{alg:secondlevel}}(T,C)=\tO(C^{\alpha}T^{1-\alpha}),
	\end{align}
	in time horizon $T$, and optimal parameter changes $C$.
	\begin{proof}
		By combining Lemma \ref{lem:A5}, \ref{lem:A4}, \ref{lem:P}, we have
		\begin{align}
			R_{A\ref{alg:secondlevel}}(T,C)=&O(\sqrt{T\log\log (T)}) +O(\sqrt{\wC T\log (T/\wC)})\\ &+O(\wC^{\alpha}T^{1-\alpha}).
		\end{align}
		Putting in the result in Lemma \ref{lem:wC} and the fact that $\alpha\leq 0.5$, $C\leq T$ concludes the proof.
	\end{proof}
\end{theorem}

Thus, with Alg. \ref{alg:secondlevel} and Theorem \ref{thm:A5}, we have successfully created a near-optimal algorithm (i.e., $\tO(C^\alpha T^{1-\alpha})$ regret) with $\log^2(T)$ computational complexity overhead.

\section{A Log-complexity Optimal Algorithm}\label{sec:lognear}
\begin{algorithm}[!t]
	\caption{Recursive Merging}\label{alg:recursive}
	{\begin{algorithmic}[1]
			\STATE Inputs $T$
			\STATE Create expert $x_{0,t}$ which is the belief of Alg. \ref{alg:belief}
			\STATE Create expert $x_{1,t}$ as follows: for $t\in\{1,\ldots,T/2\}$, it is the belief of an Alg. \ref{alg:recursive} with input $T/2$; and for $t\in\{T/2+1,\ldots,T\}$ it is the belief of another Alg. \ref{alg:recursive} with input $T/2$
			\STATE Set $\eta=\sqrt{\frac{1}{T}}$
			\STATE Initialize $p_{0,1}=1/2$ and $p_{1,1}=1/2$
			\FOR {$t=1,2,\ldots,T$}
			\STATE Draw $I$ from the distribution $p_{i,t}$
			\STATE $\xt[t]=\xt[I,t]$
			\STATE Incur loss $\lt(\xt)$
			\STATE $\tpt[i,t+1]=\pt[i,t]e^{-\eta \expect{l_t(\xt[i,t])}}$
			\STATE $\pt[i,t+1]=\dfrac{\tpt[i,t+1]}{\sum_{j=1}^2\tpt[j,t+1]}.$
			\STATE Output $\expect{l_t(x_t)}\triangleq p_{0,t}l_t(x_{0,t})+p_{1,t}\expect{l_t(x_{1,t})}$
			\ENDFOR
	\end{algorithmic}}
\end{algorithm}
In this section we develop an optimal algorithm, which has $\log(T)$ computational complexity overhead. We observe that while the first-level merging in Section \ref{sec:log2near} is required to aggregate different reset times (which also exists in the algorithm of Section \ref{sec:logsub}), the second-level merging is only for a parameter optimization scheme. Hence, if we can merge the belief in a way that this optimization is no longer required, then we can create a more efficient algorithm. We do this inherent optimization by recursive merging of the beliefs instead of the traditional parallel merging, which we name the Recursive Experts.

\subsection{The Recursive Experts}
Instead of a time interval selection scheme, our dynamic algorithm develops naturally by considering the aggregation of the static algorithm (the base algorithm) with a dynamic version recursively. In our algorithm, given a time horizon $T$, we will merge the static algorithm, which will run for $T$ times, with its dynamic counterpart, which will reset at the middle of the time horizon, i.e., $T/2$, to incorporate adaptivity. Thus, our recursive algorithm works as the following. For a given runtime $\tau$, the recursive algorithm is a mixture between the learning system given in Alg. \ref{alg:belief} that runs for $\tau$ times and the recursive algorithm itself with runtime $\tau/2$ followed by another run of the recursive algorithm with runtime $\tau/2$. 

We denote $\xt[0,t]$ as the belief of the static algorithm at time $t$, and $\xt[1,t]$ as the belief of the dynamic algorithm (the recursive mixture). We denote their selection probabilities $\pt[0,t]$ and $\pt[1,t]$ respectively. We set $x_t$ as our belief by randomly selecting $\xt[i,t]$ with the probabilities $\pt[i,t]$, and output the expected loss $\expect{l_t(x_t)}\triangleq p_{0,t}l_t(x_{0,t})+p_{1,t}\expect{l_t(x_{1,t})}$ to use for the mixture in the higher level of the recursion. We again create the probabilities $\pt[i,t]$ with the exponentiated expected losses. A description of the algorithm is given in Alg. \ref{alg:recursive}. Next, we will study its regret.

\subsection{Performance Analysis}
In this section, we analyze the regret of our algorithm. Let the regret of Alg. \ref{alg:recursive} be $R_{A\ref{alg:recursive}}(T,C)$ for a time horizon $T$ and number of changes $C$. We structure the indexing of the recursive algorithm with a binary tree splitting, i.e., the recursive algorithm indexed $(i,j)$ will be the $j^{th}$ sequential run at the $i^{th}$ level, which splits into the $(2j-1)^{th}$ and $(2j)^{th}$ runs at the $(i+1)^{th}$ level in its recursion, starting from the index $(0,1)$ at the top-level. Let $R_{A\ref{alg:recursive}}^{(i,j)}(T_i,C_{i,j})$ be the regret of the recursive algorithm indexed $(i,j)$. Then, we naturally have
\begin{align}
	R_{A\ref{alg:recursive}}(T,C)=R_{A\ref{alg:recursive}}^{(0,1)}(T_0,C_{0,1}),
\end{align}
where $T_0=T$ and $C_{0,1}=C$.

\subsubsection{Recursive Regret}
Our regret bounds will have a recursive relation as will be shown in the following two Lemmas.
\begin{lemma}\label{lem:A6c1}
	Alg. \ref{alg:recursive} indexed with $(i,j)$ will have the following regret when $C_{i,j}=1$
	\begin{align}
	R_{A\ref{alg:recursive}}^{(i,j)}(T_i,1)\leq O(\sqrt{T_i})+R_{A\ref{alg:belief}}(T_i),
	\end{align}
	where $R_{A\ref{alg:belief}}(\tau)$ is the regret of the base algorithm in $T_i$ times as in Assumption \ref{ass:RB}.
	\begin{proof}
		The proof follows from the fact that the recursive algorithm is a mixture of two experts between the base algorithm and the next level of the recursive algorithm. When using the learning parameter $\eta\sim T_i^{-\frac{1}{2}}$ in a mixture of two experts setting \cite{doubling_trick}, we incur the redundancy $O(\sqrt{T_i})$.
	\end{proof}
\end{lemma}

\begin{lemma}\label{lem:A6cC}
	Alg. \ref{alg:recursive} indexed with $(i,j)$ will have the following regret when $C_{i,j}>1$
	\begin{align}
	R_{A\ref{alg:recursive}}^{(i,j)}(T_i,C_{i,j})\leq& O(\sqrt{T_i})+R_{A\ref{alg:recursive}}^{({i+1,2j})}(T_{i+1},C_{i+1,2j})\\
	&+R_{A\ref{alg:recursive}}^{(i+1,2j-1)}(T_{i+1},C_{i+1,2j-1})
	\end{align}
	where $C_{i+1,2j-1}+C_{i+1,2j}\leq C_{i,j}+1$ and $T_{i+1}=T_i/2$.
	\begin{proof}
		The proof follows from the fact that the recursive algorithm is a mixture of two experts between the base algorithm and the deeper level of the recursive algorithm. The number of $C_{i,j}$ segments in $T_i$ times will split into $C_{i+1,2j-1}$ and $C_{i+1,2j}$ segments which happen in the former and the latter $T_i/2$ run of the recursive algorithm, respectively. Because of the split at $T_i/2$, total number of changes are increases by at most $1$.
	\end{proof}
\end{lemma}

\begin{corollary}\label{cor:Cij}
	For any level $i$, we have the following
	\begin{align}
		\sum_{j=1}^{2^i}(C_{i,j}-1)\leq C-1.
	\end{align}
	\begin{proof}
		The proof follows from Lemma \ref{lem:A6cC}; where when we move in the recursion, we have $(C_{i+1,2j-1}-1)+(C_{i+1,2j}-1)\leq C_{i,j}-1$.
	\end{proof}
\end{corollary}

\subsubsection{Mixture Redundancy}
Let $R_M^{(i,j)}(T,C)$ be the mixture redundancy we incur from $j^{th}$ run of the recursive algorithm at level $i$ for time horizon $T$ and number of segments $C$. This redundancy is the first term in the regret bounds of Lemma \ref{lem:A6c1} and \ref{lem:A6cC}. From these Lemmas, we observe that while Lemma \ref{lem:A6cC} holds, this redundancy is $O(\sqrt{T_i})$; however, when Lemma \ref{lem:A6c1} is true for $(i,j)$, the mixture redundancy of the recursion from then on becomes $0$ since the recursion stops at $(i,j)$.
\begin{lemma}\label{lem:RMC}
	The total mixture redundancy of Alg. \ref{alg:recursive} with time horizon $T$ and parameter change $C$ is
	\begin{align}
		R_M(T,C)=O(\sqrt{CT}).
	\end{align}
	\begin{proof}
	For each recursive level $i$ of the algorithm with time length $T_i$, we have a regret redundancy of $O(\sqrt{T_i})$ because of the parallel merging of two algorithms as stated in Lemma \ref{lem:A6c1} and \ref{lem:A6cC} (the first terms). Starting from the top-level, we will only move down to a lower level, if there is a change in that specific time segment of the recursive algorithm (i.e., $C_{i,j}>1$). 
	At each $i^{th}$ level, we will have $2^{i}$ number of segments (i.e., $j\in\{1,\ldots,2^i\}$) of length $2^{-i}T$ (assume $T=2^N$ for some $N$). For the total mixture redundancy of the $i^{th}$ level, $R_M^{(i)}(T,C)$, we have
	\begin{align}
		R_M^{(i)}(T,C)=\sum_{j=1}^{2^i}R_M^{(i,j)}(T,C).
	\end{align}
	
	Since the total parameter change is $C$, from Corollary \ref{cor:Cij}, we know that there can be at most $C-1$ segments with $C_{i,j}>1$, which will be split and move down in the recursion. Once we move down, we will incur the mixture redundancy of that level where the time length is $2^{-i-1}T$ as in Lemma \ref{lem:A6c1} and \ref{lem:A6cC}. If $2^{i}\leq C$, then, at worst case, we will incur redundancy from all $2^{i+1}$ segments in the next level. On the other hand, if $C\leq 2^{i}$, we will recur redundancy from at most $2C$ segments (because of the split). Thus, the redundancy incurred from $(i+1)^{th}$ level is
	\begin{align}
	R_M^{(i+1)}(T,C)\leq\min(2C,2^{i+1})O\left(\sqrt{\frac{T}{2^{i+1}}}\right),\label{RMC}
	\end{align}
	since $M_{i+1,j}$ are either $0$ or $O(\sqrt{T/2^{i+1}})$. 
	Let $2^K\leq C\leq 2^{K+1}$ for some $K$. Then, we have
	\begin{align}
		R_M(T,C)\triangleq& \sum_{i=0}^{N}R_M^{(i)}(T,C)\nonumber
		\\\leq&\sum_{i=0}^{K}O\left(\sqrt{{2^{i}T}}\right)+\sum_{i=K+1}^{N}O\left(\sqrt{\frac{4C^2T}{2^{i}}}\right)\nonumber
		\\\leq&O\left(\sqrt{2^{K+1}T}\right)+O\left(\frac{C\sqrt{T}}{\sqrt{2^{K+1}}}\right)\nonumber
		\\=&O\left(\sqrt{CT}\right),\label{eq:5}
	\end{align}
	where we used \eqref{RMC}, the sum of power series, and the bounds on $C$ successively, which concludes the proof.
	\end{proof}
\end{lemma}

\subsubsection{Regret of the Base Algorithms}
Next, we study the regret of the time segments that run Alg. \ref{alg:belief}. After the mixture, we reach the regret of these time segments by lieu of Lemma \ref{lem:A6c1}.
\begin{lemma}
	The cumulative regret resulting from Alg. \ref{alg:belief} that is used in Alg. \ref{alg:recursive} in a time segment of length $t_c$ (where the optimal in hindsight remains unchanged) is
	\begin{align}
		R_S(t_c)=O(t_c^{1-\alpha}).
	\end{align}
	\begin{proof}
		In our algorithm, each time segment $t_c$ is broken down to smaller parts of $\tau_m$ for $m\in\{1,\ldots,M\}$ for some $M$. There are two possible cases: either for some $m^*$, $\tau_{m^*}$ is singlehandedly the largest part; or for some $m^*$ and $m^*+1$, $\tau_{m^*}$ and $\tau_{m^*+1}$ are equally the largest part. In both cases, $\tau_m$ is strictly increasing for $m<m^{*}$ (otherwise we could combine the adjacent equal length parts). Similarly, $\tau_m$ is strictly decreasing for $m>m^{*}$ and $m>m^{*}+1$ respectively for the same reasons. In the first case, we see that $\tau_{m*}\leq t_c\leq 3\tau_{m^*}-2$. Similarly, in the second case, $2\tau_{m^*}=2\tau_{m^*+1}\leq t_c\leq 4\tau_{m^*}-2$. In both cases, the left and right sides of the biggest part(s) will be made of a subset of the possible part lengths (which are powers of $2$) less than $\tau_{m^*}$. Let $\tau_{m^*}=2^l$. To upper bound the regret, we may simply sum for all the part length up to $2^l$. For both cases, this is
		\begin{align}
			R_S(t_c)\leq& 2\sum_{i=0}^{l}R_{A\ref{alg:belief}}(2^i)
			\\=&2\sum_{i=0}^{l}O\left(2^{(1-\alpha)i}\right)
			\\\leq& O\left(t_c^{1-\alpha}\right),
		\end{align}
		which uses the sum of power series, the fact that $\tau_{j^*}\leq t_c$, and $\alpha\leq 0.5$ to conclude the proof.
	\end{proof}
\end{lemma}
Hence, for the $C$ segments where the optimal parameter in hindsight remains unchanged, we have the following Lemma.
\begin{lemma}\label{lem:RSC}
	The cumulative regret resulting from Alg. \ref{alg:belief} that is used in Alg. \ref{alg:recursive} in a time horizon $T$ is
	\begin{align}
		R_{SC}(T,C)=O(C^{\alpha}T^{1-\alpha}).
	\end{align}
	\begin{proof}
		The proof comes from the concavity of $R_S(\cdot)$.
	\end{proof}
\end{lemma}

\subsubsection{Total Regret of the Recursive Experts}
Thus, by summing the mixture redundancy with the regret of the base algorithms, we reach the total regret of our approach.
\begin{theorem}\label{thm:rec}
	The regret of Alg. \ref{alg:recursive} for known time horizon $T$ is
	\begin{align}
	R_{A\ref{alg:recursive}}(T,C)=O\left(C^{\alpha}T^{1-\alpha}\right).
	\end{align}
	\begin{proof} 
	By combining the results of Lemma \ref{lem:RMC}, \ref{lem:RSC}; we have
		\begin{align}
		R_{A\ref{alg:recursive}}(T,C)\triangleq&R_{MC}(T,C)+R_{SC}(T,C).
		\\=&O(\sqrt{CT})+O(C^{\alpha}T^{1-\alpha})
		\\=&O(C^{\alpha}T^{1-\alpha}),
		\end{align}
		since $\alpha\leq 0.5$, $C\leq T$, which concludes the proof.
	\end{proof}
\end{theorem}

Thus, with Alg. \ref{alg:recursive} and Theorem \ref{thm:rec}, we have succeeded in constructing an optimal algorithm (i.e., $O(C^\alpha T^{1-\alpha})$ regret) with only $\log(T)$ computational complexity overhead.

\section{Conclusion}\label{sec:conc}
In this paper, we have introduced an algorithm for general sequential learning systems that achieves adaptivity in dynamic environments, which incorporates reset periods into the learning systems and efficiently merges their beliefs.
We have developed three algorithms with varying degrees of efficiency to merge sequential learning systems to work in dynamic environments. Our initial algorithm in our development was a simple merger with log-complexity sub-optimal regret bounds, i.e., $o(T)$. Our second algorithm was smarter in the sense that it comprehensively utilized its hyper-experts derived from the base algorithm to achieve near-optimal regret bounds, i.e., $\tO(CR(T/C))$, where $R(\cdot)$ is the static regret of the base algorithm, with $\log^2(T)$ complexity. Our final algorithm, which we call the Recursive Experts was more sophisticated, where with only $\log(T)$ complexity, we were able to achieve optimal regret bounds, i.e., $O(CR(T/C))$. Hence, we have constructed an efficient strategy of merging general learning systems, which is minimax optimal up to constant factors.
	
\bibliographystyle{ieeetran}
\bibliography{double_bib}	
	
	\begin{appendices}
	\end{appendices}
	
\end{document}